\documentclass{article}

     \usepackage{arxiv}

\usepackage[utf8]{inputenc} 
\usepackage[T1]{fontenc}    
\usepackage{hyperref}       
\usepackage{url}            
\usepackage{booktabs}       
\usepackage{amsfonts}       
\usepackage{nicefrac}       
\usepackage{microtype}      
\usepackage{xcolor}         
\usepackage{amsmath}
\usepackage{graphicx}
\usepackage{tabularx}
\usepackage{threeparttable}
\usepackage[numbers]{natbib}

\title{CAVACHON: a hierarchical Variational Autoencoder to integrate multi-modal single-cell data}

\author{%
  Ping-Han Hsieh\\
  Centre for Molecular Medicine Norway (NCMM),\\
  Nordic EMBL Partnership,\\
  University of Oslo, Oslo, Norway;\\
  Department of Informatics,\\
  University of Oslo, Oslo Norway;\\
  \texttt{pinghanh@ncmm.uio.no} \\
  \And
  Ru-Xiu Hsiao\\
  Genome and Systems Biology Degree Program,\\
  National Taiwan University, Taipei, Taiwan;\\
  \texttt{r09b48004@ntu.edu.tw}
  \And
  Katalin Ferenc\\
  Centre for Molecular Medicine Norway (NCMM),\\
  Nordic EMBL Partnership,\\
  University of Oslo, Oslo, Norway;\\
  \texttt{k.t.ferenc@ncmm.uio.no} \\
  \And
  Anthony Mathelier\\
  Centre for Molecular Medicine Norway (NCMM),\\
  Nordic EMBL Partnership,\\
  University of Oslo, Oslo, Norway;\\
  Center for Bioinformatics,\\
  University of Oslo, Oslo, Norway;\\
  Department of Medical Genetics, \\
  Oslo University Hospital, Oslo, Norway;\\
  \texttt{anthony.mathelier@ncmm.uio.no} \\
  \And
  Rebekka Burkholz\\
  CISPA Helmholtz Center for Information Security,\\
  Saarland, Germany;\\
  \texttt{burkholz@cispa.de} \\
  \And
  Chien-Yu Chen\\
  Department of Biomechatronics Engineering,\\
  National Taiwan University, Taipei, Taiwan;\\
  Genome and Systems Biology Degree Program,\\
  National Taiwan University, Taipei, Taiwan;\\
  \texttt{chienyuchen@ntu.edu.tw}\\
  \And
  Geir Kjetil Sandve\\
  \;\;\;\;\;\;\;\;\;\;\;\;\;Department of Informatics,\;\;\;\;\;\;\;\;\;\;\;\;\;\\
  University of Oslo, Oslo, Norway\\
  \texttt{geirksa@ifi.uio.no} \\
  \And
  Tatiana Belova$\dagger$\\
  Centre for Molecular Medicine Norway (NCMM),\\
  Nordic EMBL Partnership,\\
  University of Oslo, Oslo, Norway;\\
  \texttt{tatiana.belova@ncmm.uio.no} \\
  \And
  Marieke L. Kuijjer$\dagger$ \\
  Centre for Molecular Medicine Norway (NCMM),\\
  Nordic EMBL Partnership,\\
  University of Oslo, Oslo, Norway;\\
  Department of Pathology\\
  Leiden University Medical Center;\\
  \texttt{marieke.kuijjer@ncmm.uio.no} \\
}

\begin{document}

\maketitle
\newpage

\begin{abstract}
Paired single-cell sequencing technologies enable the simultaneous measurement of complementary modalities of molecular data at single-cell resolution. Along with the advances in these technologies, many methods based on variational autoencoders have been developed to integrate these data. However, these methods do not explicitly incorporate prior biological relationships between the data modalities, which could significantly enhance modeling and interpretation. We propose a novel probabilistic learning framework that explicitly incorporates conditional independence relationships between multi-modal data as a directed acyclic graph using a generalized hierarchical variational autoencoder. 
We demonstrate the versatility of our framework across various applications pertinent to single-cell multi-omics data integration. These include the isolation of common and distinct information from different modalities, modality-specific differential analysis, and integrated cell clustering.
We anticipate that the proposed framework can facilitate the construction of highly flexible graphical models that can capture the complexities of biological hypotheses and unravel the connections between different biological data types, such as different modalities of paired single-cell multi-omics data. The implementation of the proposed framework can be found in the repository \url{https://github.com/kuijjerlab/CAVACHON}.
\end{abstract}

\section{Introduction}
In the past decade, the emergence of high throughput single-cell technologies has enabled the profiling of snapshots of gene and protein expression~\citep{tang2009mrna, kivioja2012counting, budnik2018scope}, chromatin accessibility~\citep{cusanovich2015multiplex, lake2018integrative, mezger2018high}, and transcription factor binding~\citep{grosselin2019high} occurring in each cell. Single-cell datasets generated with these technologies have broadened our understanding of, for example, stochastic gene expression ~\citep{islam2014quantitative}, expression dynamics during cell differentiation~\citep{cuomo2020single}, and differences in chromatin states between cells from healthy and diseased tissues~\citep{satpathy2018transcript, domcke2020human}. However, regulatory proteins cooperate to bind to specific non-coding DNA regions, known as regulatory elements, to mediate the biological processes in the cell. Using data from single-omics technologies may fail to fully capture the complex molecular mechanisms involving multiple dimensions or modalities of regulatory elements and proteins~\citep{argelaguet2021computational}.

Thanks to recent advances in single-cell multi-omics sequencing technology such as CITE-Seq~\citep{stoeckius2017simultaneous}, scNMT-Seq~\citep{clark2018scnmt}, SNARE-Seq~\citep{chen2019high}, SHARE-Seq~\citep{ma2020chromatin}, and 10X Multiome~\citep{belhocine2021single}, it is now possible to capture multiple modalities of molecular data simultaneously at single-cell resolution. However, integrating and deriving biological insights from multiple modalities is challenging, as each modality requires distinct data processing, normalization, modeling, and interpretation~\citep{argelaguet2021computational}. Various data analysis tools and computational methods have been proposed to address this. For instance, Seurat~\citep{hao2021integrated}, Harmony~\citep{korsunsky2019fast},  Scanpy~\citep{wolf2018scanpy}, and MUON~\citep{bredikhin2022muon} provide unified frameworks for quality control, data exploration, and downstream analysis of single-cell data. 
Other methods such as scMVAE~\citep{zuo2021deep}, Cobolt~\citep{gong2021cobolt}, scMM~\citep{minoura2021mixture}, and TotalVI~\citep{gayoso2021joint} integrate multiple modalities by applying variational autoencoders to embed the observed data into a joint low-dimensional latent space. However, using a joint latent space makes it difficult to interpret the model and unravel interactions between different modalities. Moreover, most of these methodologies aim at direct multi-omics data integration without integrating biological hypotheses. How to incorporate a prior biological understanding of the properties of data into the existing model remains an open question in the field. GLUE~\citep{cao2022multi} was recently proposed to incorporate a prior knowledge graph between biological features to link the generative process between different modalities. This approach requires constructing connections between the regulatory elements, regulatory proteins and genes in advance, making it difficult to apply when the understanding of the detailed underlying regulatory mechanisms is lacking. 

To address these shortcomings, we propose CAVACHON---or Cell cluster Analysis with Variational Autoencoder using Conditional Hierarchy Of latent representioN---, a novel probabilistic learning framework that incorporates the conditional independence relationships on the modality-level using a generalized variational ladder autoencoder~\citep{zhao2017learning}. This framework takes as input a single-cell mulit-omics data set and a directed acyclic graph (optional) representing the relationships between modalities. The model is sequentially trained based on the topological order in the provided graph to optimize the evidence lower bound (ELBO) of the data likelihood of each modality.

Applying our method to a SNARE-Seq dataset from the cerebral cortex of an adult mouse, we show that it can isolate common and distinct variability of gene expression and chromatin accessibility signals. Moreover, the analysis of a 10X Multiome dataset from human peripheral blood mononuclear cells (PBMCs) identifies modality-specific differential expression and the underlying regulators. We anticipate that the proposed framework will help users construct flexible graphical models that easily reflect biological hypotheses and unravel the interactions between biological data types, such as different modalities of paired single-cell multi-omics data.

\section{Methods}
\subsection{Problem Definition}
\begin{figure}[h]
  \includegraphics[width=\textwidth]{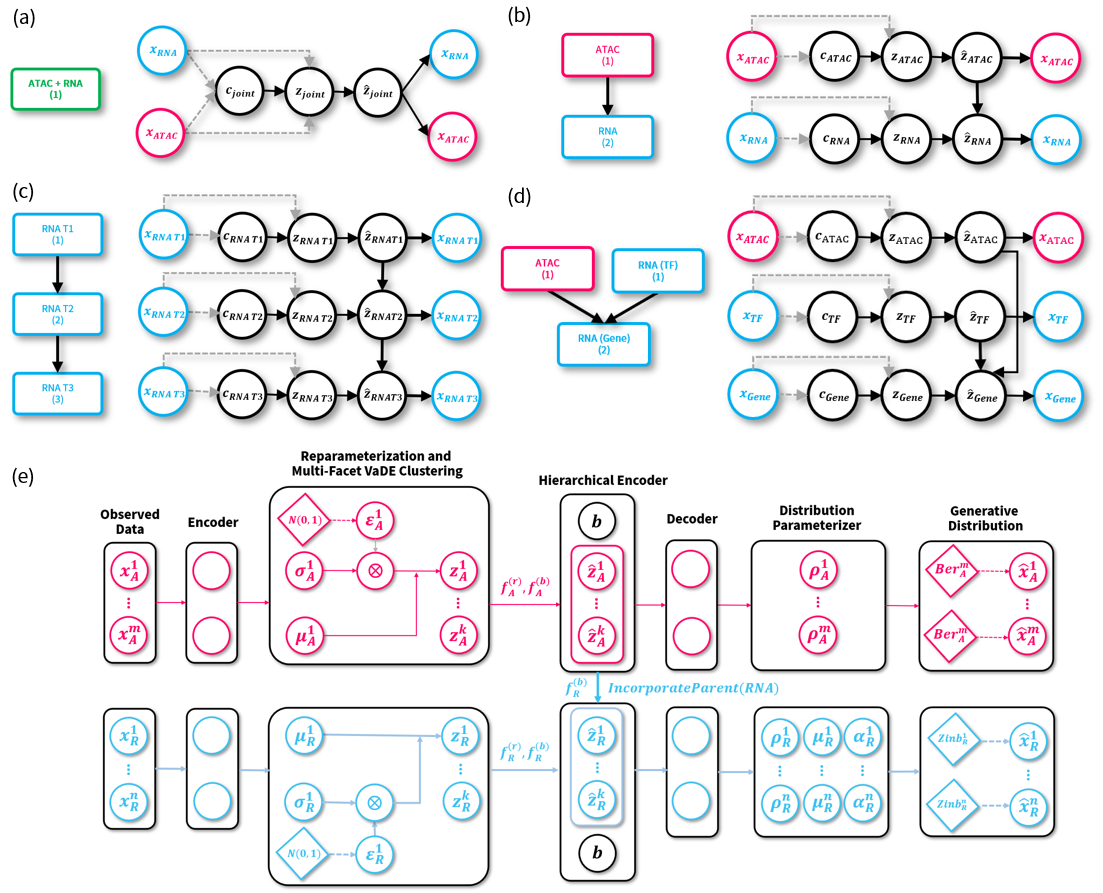}
  \centering
  \caption{Examples of relational graphs and graphical models. (a)-(d) Schematic diagrams of input graphs representing conditional independent relationships (left) between modalities of data and the graphical models created by our method (right). Note that, for simplicity, the batch information is omitted. The numbers under the names of the modalities in the rectangle boxes are the topological orders used for sequential training. The nodes in the graphical model are colored by the type of molecular assays. The dashed arrows denote the recognition model for posterior approximation, while the solid arrows denote the generative process. The figure shows examples with (a) joint learning, (b) states of chromatin accessibility that influence the expression of genes, (c) gene expression of later time points in a time series experiment that is dependent on earlier time points, and (d) states of chromatin accessibility and transcription factors that regulate gene expression. (e) The architecture of the created hierarchical variational autoencoder for the graphical model of (b).}
  \label{figure:relational_graphs}
\end{figure}

Consider $M$ multiple modalities of data with the same cell identifier in paired single-cell multi-omics data, and a directed acyclic graph $\mathcal{G}=(\mathcal{V}, \mathcal{E})$, where $\mathcal{V}=\{v_m\;|\;m=1,2...M\}$ is the set of vertices representing the data modalities, and $\mathcal{E}\subseteq \mathcal{V}\times\mathcal{V}$ is the set of unweighted directed edges specifying our prior knowledge on the conditional independence relationships between different modalities. Conditional independence relationships between different modalities are provided as follows: if a directed edge exists from modality A to modality B, then the two modalities are conditionally independent given the latent distribution of modality A (see Figure \ref{figure:relational_graphs} a-d). Practically, our probabilistic model learns the latent distribution of modality A which delineates a shared biological mechanism underlying the observed data in modalities A and B. On the other hand, the latent distribution of modality B captures the unique, independent features of modality B that remain unexplained by the shared biological mechanism. Consequently, the generative process of the observed data is learned by accounting for the provided dependencies between the latent distributions across different modalities. This is achieved through the construction and training of a generalized hierarchical variational autoencoder explicitly based on $\mathcal{G}$.

For paired single-cell multi-omics sequencing data, each modality represents a single molecular assay anchored by the barcodes of cells. The graphical prior knowledge provided as input in the method could, for example, be given by the states of chromatin accessibility that may facilitate regulation of gene expression~\citep{klemm2019chromatin} (Figure \ref{figure:relational_graphs}b). Note that although we mainly present our proposed method in the context of paired single-cell multi-omics data, it can also be applied to single-omics data derived from single cells or bulk tissues, 
e.g. time series data analysis (Figure \ref{figure:relational_graphs}c). Please refer to Figure \ref{figure:relational_graphs} for other potential use cases.

\subsection{Probabilistic Generative Model}
\label{section:model}
Inspired by the Multi-Facet Clustering Variational Autoencoders (MFCVAE) designed for multi-facet clustering in computer vision~\citep{falck2021multi}, we designed a generalized version of a variational ladder autoencoder to incorporate prior knowledge of the conditional independence relationships between biological data into a probabilistic learning framework. This approach allows multiple cluster assignments that conditionally separate the abstract features of each data modality, based on a provided relational graph. Our proposed model approximates the generative process of the observed multi-modal data explicitly based on the prior conditional independence relationships between modalities (Equation~\ref{equation:generative_model}).

\begin{equation}
\begin{aligned}
    c_m &\sim Categorical(\pi_{m}) \\
    z_m &\sim IndependentNormal(\mu(c_m), \Sigma(c_m)) \\
    \tilde{z}_m&=f^{(r)}_m(z_m; \theta^{(r)}_m)\\
    \hat{z}_m &= IncorporateParents(m)\\
    \rho_m &= f^{(d)}_m([\hat{z}_m; b]; \theta_m^{(d)})\\
    x_m &\sim Dist(\rho_m),
\end{aligned}
\label{equation:generative_model}
\end{equation}

where $c_m$ is the cluster assignment for the mixture of independent Gaussian distributions with the corresponding mean $\mu(c_m)$ and standard deviation $\Sigma(c_m)$ and $b$ is the batch information that needs to be corrected for the cells. We correct for batch effects by conditioning the decoder on $b$~\citep{lopez2018deep, ashuach2022peakvi}. $z_m$ is the latent representation of modality $m$. $IncorporateParents$ is a function that incorporates the latent representations of the parents of the $m$-th modality $\mathcal{P}(m)=\{p|(p,m)\in\mathcal{E}\}$ into the posterior approximation considering the prior conditional independence relationships specified in $\mathcal{G}$. Here, we concatenate the latent representations of the parent modalities and condition the posterior distribution on them as follows:

\begin{equation}
IncorporateParents(m)=f^{(b)}_m([\hat{z}_\mathcal{P};\tilde{z}_m]; \theta^{(b)}_m),
\label{equation:z_hat}
\end{equation}

where $f^{(b)}_m$ and $f^{(r)}_m$ are linear transformations with parameters $\theta^{(b)}_m$ and $\theta^{(r)}_m$. $f^{(d)}_m$ is the decoder neural network used to parameterize the specified data distribution. $\rho_m$ is the set of parameters of user-defined data distributions $Dist$ used to compute the data likelihood. A more detailed description of the framework and neural architecture of $f^{(b)}_m$, $f^{(r)}_m$, $f^{(e)}_m$ and $f^{(d)}_m$ is shown in Appendix Table~\ref{table:architecturee}. The generative process of $m$-th modality can be structured as $p_\theta^{(m)}(x_m)=p_\theta^{(m)}(x_m|z_m,z_{\mathcal{A}(m)},b)p_\theta^{(m)}(z_m|c_m)p_\theta^{(m)}(c_m)$, where $\mathcal{A}(m)$ is the set of ancestors of modality $m$ (Appendix~\ref{section_a:elbo}). 

The specifications of our proposed probabilistic learning framework's implementation are outlined in Appendix~\ref{section_a:implementation}. To train the neural network, we utilize \textit{progressive training} to acquire disentangled representations across layers within the hierarchical variational autoencoder. Further insights into the optimization procedure are provided in Appendix~\ref{section_a:training_strategy}.

\subsection{Contribution of Modalities to Differential Gene Expression}
\label{section_a:differential_analysis}
We implemented a Bayesian approach to perform differential analysis between clusters of cells. To identify differentially expressed genes or changes in chromatin states, we compute the Bayesian factor of two hypotheses $\mathcal{H}_1$ and $\mathcal{H}_2$ (see Equation~\ref{equation:differential}) following previous research~\citep{lopez2018deep}. Specifically, consider two groups of cells $\mathcal{I}, \mathcal{J}$ and the pairs from the two groups $(i, j)\subseteq \mathcal{I}\times\mathcal{J}$, where $i\in\mathcal{I}, j\in\mathcal{J}$. The Bayesian factor $K$ to evaluate whether the modality $x_m$ is differentially active between groups $\mathcal{I}$ and $\mathcal{J}$ is computed as:

\begin{equation}
\begin{aligned}
\mathcal{H}_1:\mathbb{E}_\mathcal{I\times J}[p_\theta(x_m^{(i)}&|z_\mathcal{V}^{(i)}, b_m^{(i)})]>\mathbb{E}_\mathcal{I\times J}[p_\theta(x_m^{(j)}|z_\mathcal{V}^{(j)}, b_m^{(j)})] \\
\mathcal{H}_2:\mathbb{E}_\mathcal{I\times J}[p_\theta(x_m^{(i)}&|z_\mathcal{V}^{(i)}, b_m^{(i)})]\leq\mathbb{E}_\mathcal{I\times J}[p_\theta(x_m^{(j)}|z_\mathcal{V}^{(j)}, b_m^{(j)})] \\
K=&\log\frac{p(\mathcal{H}_1|x_m^{(i)}, x_m^{(j)})}{p(\mathcal{H}_2|x_m^{(i)}, x_m^{(j)})}
\end{aligned}
\label{equation:differential}
\end{equation}

To derive the Bayesian factor $K$, we use naive Monte Carlo sampling to approximate $\mathbb{E}_\mathcal{I\times J}[p_\theta(x_m^{(i)}|z_\mathcal{V}^{(i)}, b_m^{(i)})]$ and $\mathbb{E}_\mathcal{I\times J}[p_\theta(x_m^{(j)}|z_\mathcal{V}^{(j)}, b_m^{(j)})]$. As shown in previous research~\citep{lopez2018deep}, Bayesian differential analysis allows our model to identify genes (or genomic regions) that are consistently differentially active across batches while correcting for biases stemming from variations in the number of cells. 
Using the identified differentially expressed genes (or changes in chromatin states), gene set enrichment analysis (GSEA)~\citep{subramanian2005gene} can be performed to identify enriched pathways or biological functions~\citep{fang2022gene}. 

Since our method approximates the generative process by considering the dependencies between modalities, it becomes feasible to decompose the result of Bayesian differential analysis. To achieve this, our method creates a chimeric molecular profile by integrating various modalities in different groups of cells. 
For instance, by using the latent representation of chromatin accessibility from cell type A with that of gene expression from cell type B, our generative model is capable of creating a chimeric cell with a molecular profile that captures the chromatin accessibility landscape of cell type A and the gene expression profile of cell type B (Figure ~\ref{figure:composite_deg}a).

Subsequently, differential analysis can be performed to identify genes that exhibit changes in expression patterns between cell types A and B, as well as between cell type A and the chimeric cell.
By comparing the disparities between the results from the two differential analyses, we can deconstruct the findings, identifying the changes in gene expression driven by the modality of interest (Figure ~\ref{figure:composite_deg}b).

Here, we describe the quantitative computation of the contribution of modality $m$ to the observed changes in the molecular signal $g$, denoted as $C(m, g)$. The chimeric molecular profile can be created by substituting the latent representation of modality $m$ from group $\mathcal{I}$ with the latent representation from another group $\mathcal{J}$. We compute the changes of the observed molecular signal as the contribution of modality $m$ in the generative process of the molecular signal $g$:

\begin{equation}
\begin{aligned}
C(m, g)=\mathbb{E}_\mathcal{I\times J}
[p_\theta(x_{g}^{(i)}|z_{\mathcal{V} \setminus {m}}^{(i)}, z_{m}^{(j)}, b_m^{(i)})]-\mathbb{E}_\mathcal{I\times J}
[p_\theta(x_{g}^{(i)}|z_\mathcal{V}^{(i)}, b_m^{(i)})]
\end{aligned}
\label{equation:differential_contribution}
\end{equation}

Another advantage of our method is its ability to construct intermediate stages between two groups of cells by interpolating the latent representation between them and approximating changes for the molecular signal of interest using the generative process. This capability is valuable for analyzing time-series data or comprehending cell differentiation processes.

\subsection{Datasets}
\label{section:dataset}
For the analyses presented in this study, we used SNARE-Seq data from the cerebral cortex of an adult mouse (accession number GSE126074)~\citep{chen2019high} and 10X Multiome data from human peripheral blood mononuclear cells (10k PBMCs)~\cite{genomicspbmc}. We downloaded the preprocessed data from the repository of GLUE (\href{https://scglue.readthedocs.io/en/latest/data.html}{https://scglue.readthedocs.io/en/latest/data.html}, downloaded on June 7th,  2023)~\cite{cao2022multi}. These datasets include chromatin state and gene expression modalities in the same cell. For both datasets, we binarized the chromatin accessibility data and used the Bernoulli distribution to model the data distribution. We retained the top 100,000 and 25,000 most variable peaks of the chromatin accessibility for SNARE-Seq and 10k PBMCs in the downstream analysis using the \textit{highly\_variable\_genes} function from Scanpy~\cite{wolf2018scanpy}. By default, we used Bernoulli distribution to model the chromatin accessibility data and zero-inflated negative binomial distribution to model the gene expression data, following previous studies~\cite {lopez2018deep, ashuach2022peakvi}. 

\section{Results}
\subsection{Isolating Common and Distinct Information of Modalities}
\label{isolate_common_and_distinct}
We apply our proposed probabilistic learning framework to isolate common and distinct information across various modalities using the SNARE-Seq data from the mouse cerebral cortex~\citep{chen2019high}. This dataset comprises paired chromatin accessibility states and gene expression profiles obtained from individual cells (see Method~\ref{section:dataset} for more details). 

To effectively isolate distinct information from a target modality, the latent representation corresponding to the target modality must encapsulate solely the unique information pertinent to that modality, excluding irrelevant information from other modalities. To achieve this, the latent representation of the target modality should be conditioned upon the latent representations of other modalities.

\begin{figure}[bth]
  \includegraphics[width=\textwidth]{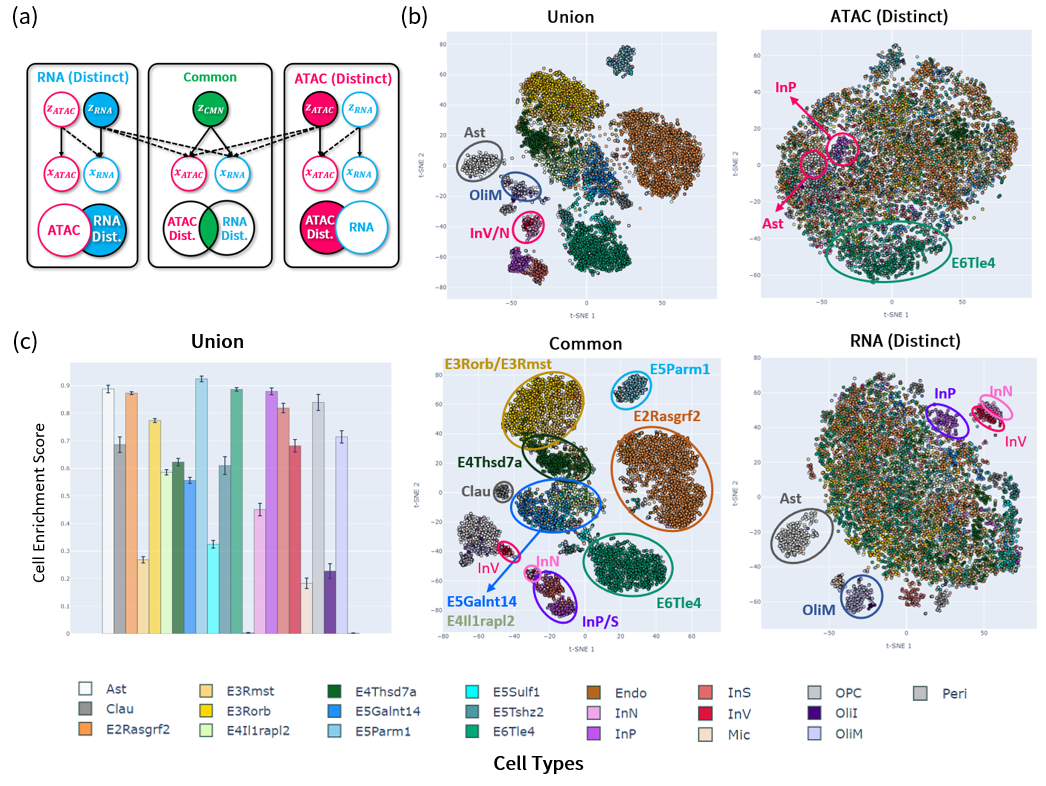}
  \centering
  \caption{Conditional latent representations for SNARE-Seq cerebral cortex of an adult mouse dataset. (a) The graphical model used to isolate the common and distinct representation between modalities. (b) t-SNE representation of the posterior mean of each modality's common and distinct latent representation. (c) Cell enrichment scores for each cell type (with 50-nearest neighbors). The cell enrichment score is the proportion of neighboring cells that share the same cell type, normalized by the number of cells in each cell type. The data points are coloured by the annotated cell types from the original SNARE-Seq study. For clarity, distinct cell types uniquely identified in each modality are annotated and highlighted with circles.}
  \label{figure:conditional_latent}
\end{figure}

For instance, utilizing the SNARE-Seq dataset, to extract the distinct representation for the RNA modality, the model architecture should be designed such that the RNA modality is conditioned upon the ATAC modality (refer to Figure~\ref{figure:conditional_latent}a, RNA Distinct). Similarly, this strategy should be employed to extract the distinct representation of the ATAC modality (refer to Figure~\ref{figure:conditional_latent}a, ATAC Distinct). Finally, the common representation can be extracted by conditioning upon the distinct components of both RNA and ATAC modalities (refer to Figure~\ref{figure:conditional_latent}a, Common). Note that this strategy can be employed for more than two modalities.

Using the cell enrichment score defined following the previous study by Lin et al., ~\citep{lin2023quantifying}, we found that CAVACHON distinguishes different cell types when considering all observed modalities, denoted as Union. (Figure ~\ref{figure:conditional_latent}c) In addition, CAVACHON is capable of isolating common and distinct information of different modalities. Notably, while most of the cell type-specific information can be captured by the common representation of RNA and ATAC modalities, the RNA modality distinctly delineates the genetic heterogeneity between astrocytes (Ast) and mature oligodendrocytes (OliM) (Figure ~\ref{figure:conditional_latent}b). 
This observation might suggest that differences in chromatin accessibility states are not the only driver of differential gene expression between these two cell types.


\subsection{Contribution of Modalities to Differential Gene Expression}
\label{regulatory_analysis_deg}
\begin{figure}[!b]
  \includegraphics[width=\textwidth]{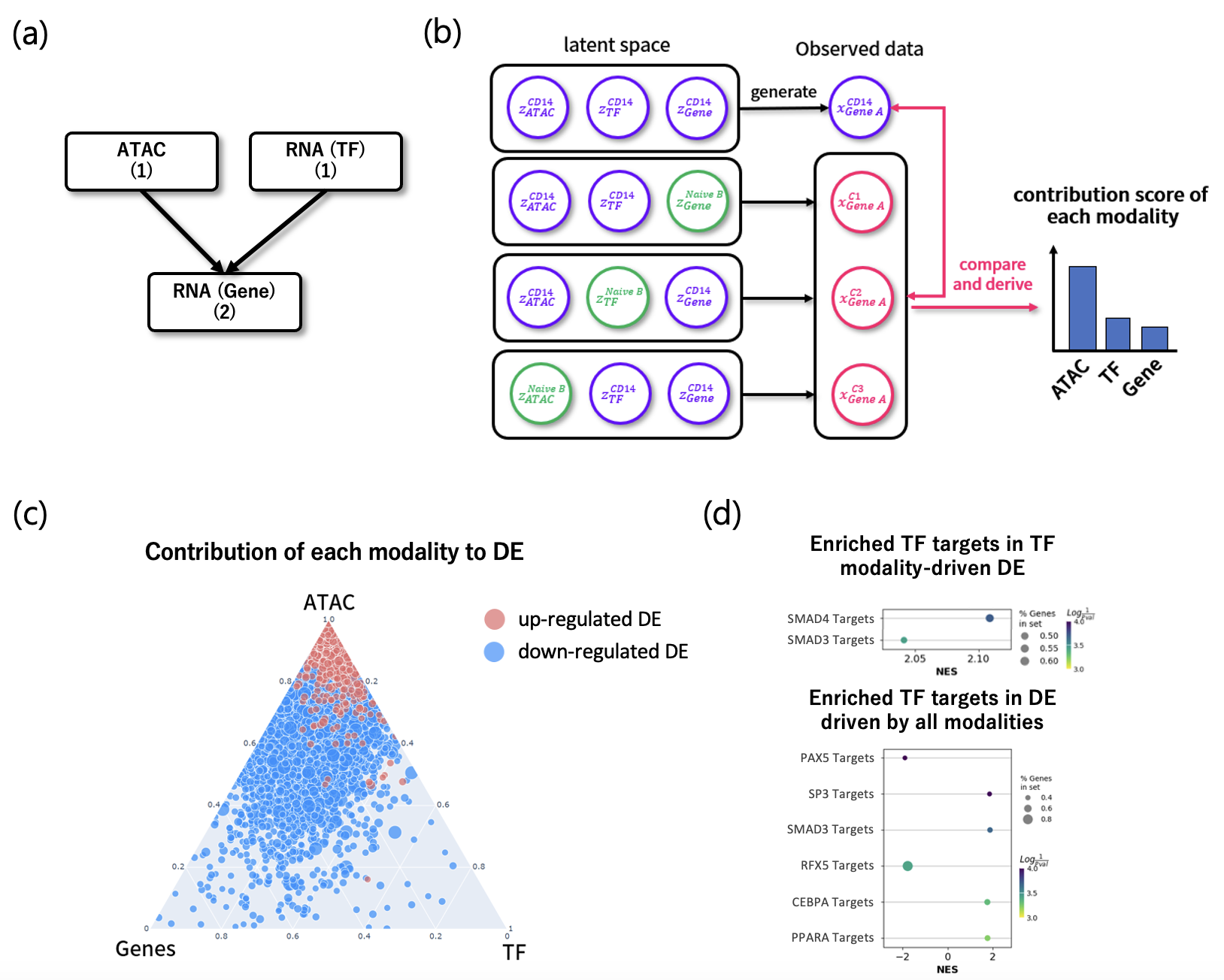}
  \centering
  \caption{(a) The graphical model used to analyze the contribution of modalities to differential gene expression in 10k PBMCs dataset. (b) The schematic diagram illustrates the computation of the contribution score for each modality. The latent distributions from CD14 monocytes are represented in purple, while those from naive B cells are represented in green. The contribution score for each modality is calculated by quantifying the change in gene expression upon substituting the latent distribution of the targeted modality. (c) The contribution score of each modality to the differentially expressed genes based on the compositional Bayesian analysis of CD14 monocytes compared to naive B cells. Each point on the ternary plot denotes a unique differential expression profile, with its position indicating the contribution score of each modality to the changes in gene expression. The closer a point is to the vertex of the plot, the greater the contribution score of the corresponding modality. The size of the points reflects the fold-change of the expression between the two conditions, and the color whether the gene is up- or down-regulated in CD14 monocytes, compared to naive B cells. (d) The enrichment of TRRUST gene sets based on the differential expression driven by transcription factor modality and all modalities. TRRUST geneset contains curated transcriptional regulatory relationships between the transcription factors and their target genes. The enrichment of a specific transcription factor indicates that most of the target genes regulated by that transcription factor exhibit a distinct profile. (NES: normalized enrichment score)}
  \label{figure:10x_degs}
\end{figure}
After applying CAVACHON to a hierarchical model including two modalities of molecular data types, we decided to present another application of the method to characterize the contribution of each modality to the observed differentially expressed genes. We analyzed the 10k PBMCs dataset, which consists of paired single-cell chromatin accessibility and gene expression data for a diverse population of immune cells found in the bloodstream (see Method~\ref{section:dataset} for more details). We use the fact that chromatin accessibility states and transcription factors influence gene expression regulation (Figure~\ref{figure:10x_degs}a). Accordingly, our probabilistic framework creates the generative process of gene expression dependent on the expression of transcription factors and chromatin accessibility. Notably, the activity of transcription factors and chromatin accessibility are conditionally independent. We observe from the conditional latent representations that chromatin accessibility captures most of the heterogeneity between cell types, while the expression of transcription factors captures the heterogeneity between their major lineages (B cells, CD8/CD4 T lymphocytes and CD14/CD16 monocytes, Figure ~\ref{figure_a:10x_latent}). This is not surprising---cell types are often characterized by widely different chromatin states~\cite{zhang2021single}, while transcription factors are generally only lowly expressed and need to be translated into protein and often undergo post-translational modifications before they become active regulators of gene expression.

CAVACHON can decompose the Bayesian differential expressed genes analysis to identify the contribution of each modality to the differentially expressed genes (Method~\ref{section_a:differential_analysis}). We specifically focus on the differentially expressed genes between CD14-positive monocytes (CD14 Mono) and naive B cells (Naive B) as they show two distinguishable clusters in the latent representations of the transcription factors, suggesting that transcription factor expression levels can partly explain the heterogeneity. 
To understand the contribution of each modality in driving differential gene expression, we computed their contribution scores (see Methods ~\ref{section_a:differential_analysis}, Figure~\ref{figure:10x_degs}b for more details). Aligned with the observation above, the majority of the differentially expressed genes are influenced by the states of chromatin accessibility, and not by the expression of transcription factors (Figure ~\ref{figure:10x_degs}c).

Next, we set out to identify the genes for which changes in overall transcription factor expression levels can explain their differential expression. 
We conducted the gene set enrichment analysis on the results from the differentially expressed genes driven by the modality of transcription factors and all modalities to identify potential upstream regulators of the differential expression. We used gene sets from the ``Transcriptional Regulatory Relationships Unraveled by Sentence-based Text mining'' (TRRUST) database, which contains curated transcriptional regulatory relationships between transcription factors and their target genes~\citep{han2015trrust}. This revealed that differences in the overall transcription factor expression levels between CD14 monocytes and naive B cells lead to changes in the expression of the target genes of SMAD3 and SMAD4 (Figure ~\ref{figure:10x_degs}d). Notably, SMAD3 and SMAD4 have been shown to act as transcriptional regulators of TGFb-induced IgA switching in B cells~\citep{park2001smad3}. Interestingly, these differences are driven by changes in the \emph{overall} expression of transcription factors as captured by the single-cell data, and not specifically by the expression levels of SMAD3 or SMAD4 themselves.  This may be due to the complexity of gene regulation [cite], as well as the fact that a transcription factor's activity does not necessarily depend on its own expression levels~\citep{sonawane2017understanding, ricci2021decoupling}.
Thus, our model captures differential regulation that may not be identified by focusing solely on the expression levels of specific transcription factors.

\subsection{Integrated Unsupervised Multi-facet Clustering}
The analysis described in Result~\ref{regulatory_analysis_deg} relies on the availability of cell type annotations for each cell. However, in scenarios where this information is lacking, \textit{unsupervised multi-facet clustering} can be used to identify groups of cells with similar molecular profiles. Our proposed framework seamlessly integrates unsupervised multi-facet clustering into the approximated generative process and thus ensures that the derived latent representations reflect the inherent clusters present in the observed data (see Appendix ~\ref{section_a:clustering}). Moreover, multi-facet clustering also considers the dependency between modalities, reflecting the heterogeneity across different cells explained by each modality~\cite{falck2021multi}. It is worth noting that the clustering strategy for each modality is learned during the training process. Therefore, performing clustering separately after training the model is not required. This reduces the computational time and facilitates the clustering of new cells onto an existing latent space without changing the clustering assignment of the original cells, which may happen when performing a separate clustering process.

\begin{figure}[h]
  \includegraphics[width=\textwidth]{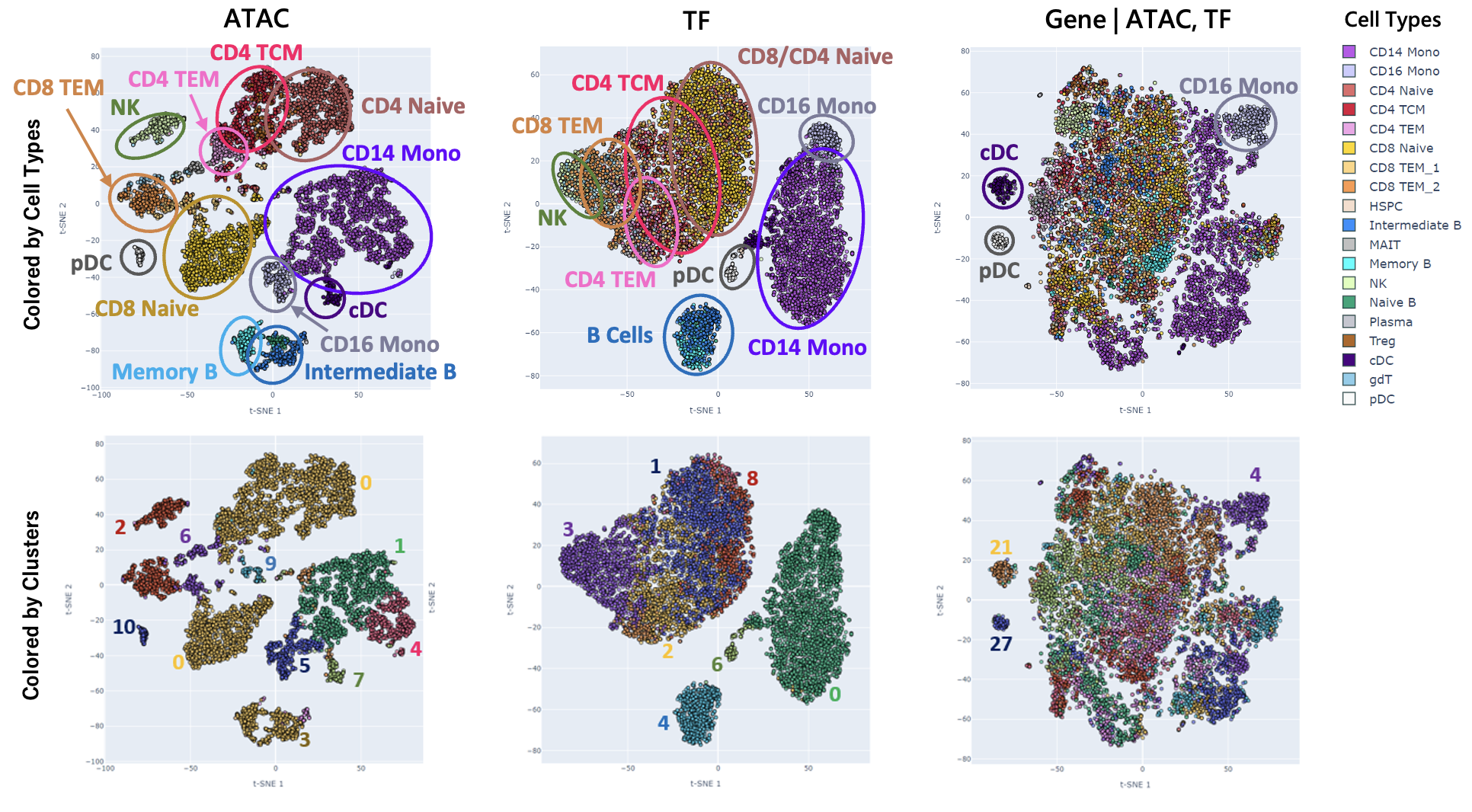}
  \centering
  \caption{Multi-facet clustering of the 10k PBMCs dataset. The top panel illustrates the posterior mean of the latent representations, colored by cell types, while the bottom panel illustrates the posterior mean of the latent representations, colored by multi-facet cluster assignments. From left to right, the panels represent the posterior mean of the latent representation of chromatin accessibility, transcription factor expression, and the expression of other genes, conditioned on the latent representations of chromatin accessibility and transcription factor expression, respectively.} 
  \label{figure_a:clustering_result}
\end{figure}

Indeed, our model can learn clustering assignments conditionally in an unsupervised way for different modalities (Figure ~\ref{figure_a:clustering_result}). Specifically, for the clustering based on chromatin accessibility only, our model captures the clusters of CD8/CD4 T lymphocytes (cluster 0), CD14-positive monocytes (cluster 1 and 4), B cells (cluster 3), plasmacytoid dendritic cells (cluster 10), and natural killer cells and CD8 effector memory T cells (cluster 2), which are both reported to show transcriptional poising of chromatin in immune response~\cite{turner2021cd8+}. For the clustering based solely on the expression of transcription factors, our model captures the different transition states of CD8/CD4 T lymphocytes (clusters 8, 1, 2 to 3 reflect the transition states of differentiation from naive cells, naïve central memory cells to effector memory cells) and CD14/CD16 monocytes (cluster 0). For the clustering based on the gene expression conditioned on both chromatin accessibility and transcription factors, our model captures CD16 monocytes (cluster 4), conventional dendritic cells (cluster 21) and plasmacytoid dendritic cells (cluster 27). 

These results indicate that CAVACHON can identify modalities driving differential gene expression and their contribution, and can isolate modality-specific heterogeneity that defines cellular phenotypes. This allows for differential analysis between clusters that considers the contribution of each modality and the characterization of cell subtypes.

\section{Discussion}
We propose a novel probabilistic learning framework to incorporate a prior biological understanding of the relationships between different data modalities in a directed acyclic graph. This framework creates a generalized hierarchical variational autoencoder that models the generative process of the observed data. The main distinction between our proposed method and previously published approaches is that the generative process modeled for each modality only uses the latent distribution of the ancestor modality defined in the graph. This allows for the explicit incorporation of prior knowledge into the generative model.

Most published methods that incorporate a variational autoencoder to integrate multi-modal data apply either the product-of-expert (PoE)~\citep{wu2018multimodal, zuo2021deep, gong2021cobolt} 
or (additive) mixture-of-expert (MoE)~\citep{shi2019variational, minoura2021mixture} factorization to approximate the posterior. 
Here, we mainly described the posterior approximation of our created model in the context of PoE. However, as previous studies suggested, MoE factorization may be more suitable for multi-modal data integration~\citep{shi2019variational}. In addition, the Gaussian posterior may limit the expressiveness of the generative model. To address this, in future studies, one could further exploit the hierarchical structure of variational autoencoder and approximate $L$ layers of posterior distributions for each modality. 
Alternatively, normalizing flows~\citep{rezende2015variational, su2018f} or diffusion models could be incorporated~\citep{rombach2022high} to approximate tighter lower bounds. We leave the exploration and evaluation of different strategies to approximate the posterior for future research.

The recently published method GLUE~\citep{cao2022multi} incorporates a prior knowledge graph representing regulatory interactions between features in different molecular assays to integrate information between modalities. The prior knowledge graph is then used to link latent representations between modalities. Although this approach shows much potential for integrative regulatory inference, it may be demanding to construct the prior knowledge graph, since our understanding of the interactions between regulatory elements is often insufficient. Our method complements GLUE by including a graph representing the dependency between different modalities. Thus, our model can identify clusters of cells that show non-canonical regulatory patterns. This is particularly useful when the understanding of a detailed regulatory network is lacking, for instance, in cancer data analysis. Unlike GLUE which utilizes neural networks to infer potential relationships between the regulators and their targets, our approach estimates and isolates the contribution of each modality to the differential gene expression. Subsequently, we identify potential regulators through gene set enrichment analysis. This enables rapid exploration of different biological hypotheses supported by previous studies. However, it is important to note that this strategy precludes the discovery of novel regulatory interactions.


Finally, our method can integrate paired single-cell multi-omics datasets in different applications. Specifically, it can isolate common and distinct heterogeneity observed in different modalities. In addition, our method performs multi-facet clustering seamlessly during model training. We provide functions for Bayesian differential analysis, considering the influence of each modality. Although we mainly describe our method in the context of single-cell data integration, it can be applied to more general settings where the provided dependency is based on modality level (e.g., time dependency in time-series data analysis). We foresee that our method will be applied to integrate other multi-omics datasets. 


\bibliography{references}
\newpage

\appendix

\section*{Appendix}
\setcounter{section}{0}
\setcounter{figure}{0}
\setcounter{equation}{0}
\renewcommand{\thefigure}{S\arabic{figure}}
\renewcommand{\thetable}{S\arabic{table}}
\renewcommand{\theequation}{S\arabic{equation}}
\section{Decomposition of ELBO}
\label{section_a:elbo}
Consider $M$ multiple modalities of data with the same anchor and a directed acyclic graph $\mathcal{G}=(\mathcal{V}, \mathcal{E})$, where $\mathcal{V}=\{v_m\;|\;m=1,2...M\}$ is the set of vertices representing the data modalities and $\mathcal{E}\subseteq \mathcal{V}\times\mathcal{V}$ the set of unweighted directed edges specifying prior knowledge on the conditional independent relationships between the different modalities. We denote a directed walk from vertex $v_i$ to $v_j$ as $w(i,j)$ and the set of all directed walks in the graph as $\mathcal{W}$. The set of ancestors of a vertex $j$ can then be defined as $\mathcal{A}(j)=\{i|w(i,j)\in \mathcal{W}\}$. The evidence lower bound (ELBO) of the conditional data likelihood can be derived as follows:

\begin{equation}
\begin{aligned}
\log{p(x_\mathcal{V}|b)}&=\mathbb{E}_{q_\phi(z_\mathcal{V},c_\mathcal{V}|x_\mathcal{V}, b)}[\log(\frac{p_\theta(x_\mathcal{V},z_\mathcal{V},c_\mathcal{V} | b)}{q_\phi(z_\mathcal{V},c_\mathcal{V}|x_\mathcal{V}, b)})]+\mathbb{E}_{q_\phi(z_\mathcal{V},c_\mathcal{V}|x_\mathcal{V}, b)}[\log(\frac{q_\phi(z_\mathcal{V},c_\mathcal{V}|x_\mathcal{V}, b)}{p_\theta(z_\mathcal{V},c_\mathcal{V}|x_\mathcal{V}, b)})]\\
&=\mathbb{E}_{q_\phi(z_\mathcal{V},c_\mathcal{V}|x_\mathcal{V}, b)}[\log(\frac{p_\theta(x_\mathcal{V},z_\mathcal{V},c_\mathcal{V} | b)}{q_\phi(z_\mathcal{V},c_\mathcal{V}|x_\mathcal{V}, b)})]+D_{\text{KL}}(q_\phi(z_\mathcal{V},c_\mathcal{V}|x_\mathcal{V}, b)||p_\theta(z_\mathcal{V},c_\mathcal{V}|x_\mathcal{V}, b))\\
&\geq\mathbb{E}_{q_\phi(z_\mathcal{V},c_\mathcal{V}|x_\mathcal{V}, b)}[\log(\frac{p_\theta(x_\mathcal{V},z_\mathcal{V},c_\mathcal{V} | b)}{q_\phi(z_\mathcal{V},c_\mathcal{V}|x_\mathcal{V}, b)})]=\mathcal{L}(x_\mathcal{V}, b;\theta,\phi)\\
\end{aligned}
\label{equation:appendix_elbo}
\end{equation}

, where $q_\phi(z_\mathcal{V}, c_\mathcal{V}|x_\mathcal{V}, b)$ are parameterized by encoder neural networks $f^{(e)}_\mathcal{V}$. Assuming independence for priors across modalities, the generative model (Equation ~\ref{equation:generative_model}) can be structured as:

\begin{equation}
\begin{aligned}
p_\theta(x_\mathcal{V}, z_\mathcal{V}, c_\mathcal{V}, b)&=p_\theta(b)\prod_{v\in\mathcal{V}}p_\theta^{(v)}(x_v|z_v,z_{\mathcal{A}(v)},b)p_\theta^{(v)}(z_v|c_v)p_\theta^{(v)}(c_v)\\
&=p_\theta(b)\prod_{v\in\mathcal{V}}p_\theta^{(v)}(x_v|z_v,z_{\mathcal{A}(v)},b)p_\theta^{(v)}(c_v|z_v)p_\theta^{(v)}(z_v)\\
p_\theta(x_\mathcal{V}, z_\mathcal{V}, c_\mathcal{V}| b)&=\prod_{v\in\mathcal{V}}p_\theta^{(v)}(x_v|z_v,z_{\mathcal{A}(v)},b)p_\theta^{(v)}(c_v|z_v)p_\theta^{(v)}(z_v)
\end{aligned}
\label{equation:generative_model_decomposed}
\end{equation}

Assuming conditional independence between the posterior of the latent representation ($z_\mathcal{V}$) and cluster assignment ($c_\mathcal{V}$) given the observed data ($x_\mathcal{V}$) and the batch ($b$), the posterior distribution can be further decomposed:
\begin{equation}
\begin{aligned}
q_\phi(z_\mathcal{V},c_\mathcal{V}|x_\mathcal{V},b)&=q_\phi(z_\mathcal{V}|x_\mathcal{V}, b)q_\phi(c_\mathcal{V}|x_\mathcal{V}, b)\\
&=\prod_{v\in\mathcal{V}}q_\phi^{(v)}(z_v|x_v, b)q_\phi^{(v)}(c_v|x_v, b)
\end{aligned}
\label{equation:posterior_decomposed}
\end{equation}

. Substituting Equation \ref{equation:generative_model_decomposed} and \ref{equation:posterior_decomposed} back to Equation (\ref{equation:appendix_elbo}), we can rewrite the ELBO as:

\begin{equation}
\begin{aligned}
\mathbb{E}_{q_\phi(z_\mathcal{V},c_\mathcal{V}|x_\mathcal{V}, b)}&[\log(\frac{\prod_{v\in\mathcal{V}}p_\theta^{(v)}(x_v|z_v,z_{\mathcal{A}(v)},b)p_\theta^{(v)}(c_v|z_v)p_\theta^{(v)}(z_v)}{\prod_{v\in\mathcal{V}}q_\phi^{(v)}(z_v|x_v, b)q_\phi^{(v)}(c_v|x_v, b)})]\\
=&\mathbb{E}_{q_\phi(z_\mathcal{V},c_\mathcal{V}|x_\mathcal{V}, b)}[\sum_{v\in\mathcal{V}}\log(p_\theta^{(v)}(x_v|z_v,z_{\mathcal{A}(v)},b))]-\\
&\mathbb{E}_{q_\phi(c_\mathcal{V}|x_\mathcal{V}, b)}D_{KL}(q_\phi(z_\mathcal{V}|x_\mathcal{V}, b) || p_\theta(z_\mathcal{V}))-\\
&\mathbb{E}_{q_\phi(z_\mathcal{V}|x_\mathcal{V}, b)}D_{KL}(q_\phi(c_\mathcal{V}|x_\mathcal{V}, b) || p_\theta(c_\mathcal{V}|z_\mathcal{V}))\\
\end{aligned}
\end{equation}


\section{Multi-facet Clustering}
\label{section_a:clustering}
To identify the optimal cluster assignment for the posterior distribution $q_\phi(c_\mathcal{V}|x_\mathcal{V}, b)$, we applied the single-facet VaDE trick~\citep{falck2021multi, jiang2016variational} to each vertex to derive the optimal posterior approximation $q_\phi^*(c_\mathcal{V}|x_\mathcal{V}, b)$ that minimizes $D_{KL}(q_\phi(c_\mathcal{V}|x_\mathcal{V}, b) || p_\theta(c_\mathcal{V}|z_\mathcal{V}))$:

\begin{equation}
\begin{aligned}
&\underset{q_\phi(c_\mathcal{V}|x_\mathcal{V}, b)}{\text{argmin}}D_{KL}(q_\phi(c_\mathcal{V}|x_\mathcal{V}, b) || p_\theta(c_\mathcal{V}|z_\mathcal{V}))=\frac{\exp{(\mathbb{E}_{q_\phi(z_\mathcal{V}|x_\mathcal{V}, b)}[\log{p(c_\mathcal{V}|z_\mathcal{V})}])}}{Z(q_\phi(z_\mathcal{V}|x_\mathcal{V}, b))} \\
&, Z(q_\phi(z_\mathcal{V}|x_\mathcal{V}, b))=\sum\limits_{k=1}^K\exp{(\mathbb{E}_{q_\phi(z_\mathcal{V}|x_\mathcal{V}, b)}[\sum_{v\in\mathcal{V}}\log{p(c_v=k|z_v)}])}
\end{aligned}
\end{equation}
, where $K$ is the number of components in the mixture of independent Gaussian priors.

\section{Implementation Details}
\label{section_a:implementation}
\begin{figure}[h]
  \includegraphics[width=\textwidth]{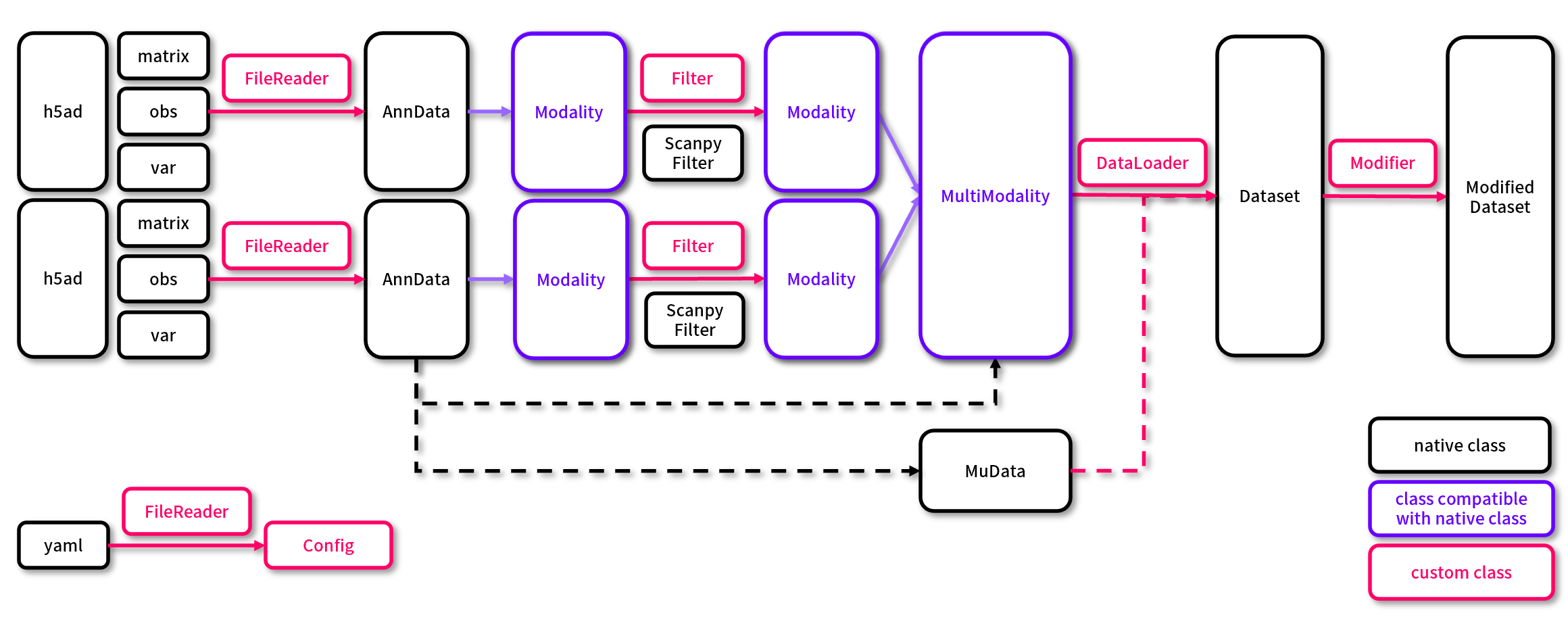}
  \centering
  \caption{Data flow with the our proposed framework. The framework takes data matrix, cell annotation, and feature annotation (or alternatively h5ad files) as inputs, constructs Modality objects (which is compatible with AnnData), and merges these into a MultiModality object (which is compatible with MuData). Finally, a Tensorflow Dataset used to load the data into the model is created from the MultiModality object.}
  \label{figure:datastructure_inputs}
\end{figure}

\begin{figure}[h]
  \includegraphics[width=\textwidth]{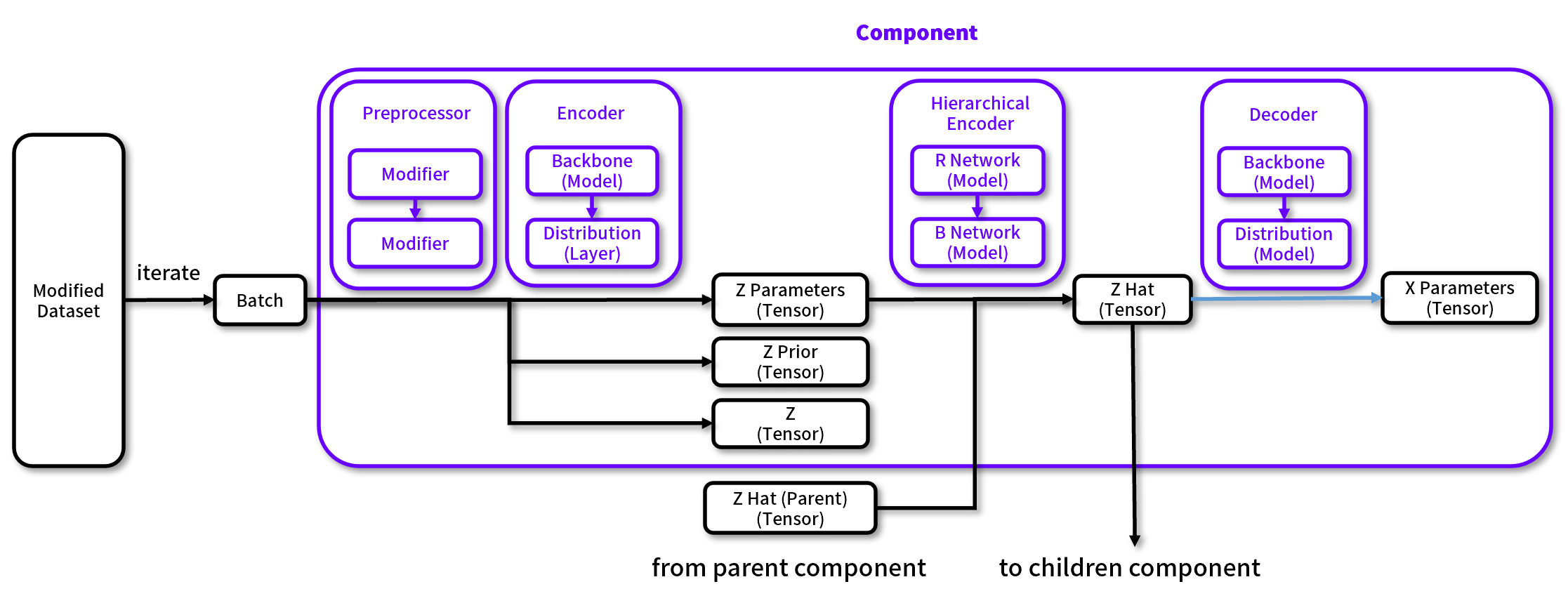}
  \centering
  \caption{Modules in each component. Every base module, including Preprocessor, Encoder, Hierarchical Encoder and Decoder, can be inherited and extended to create custom modules.}
  \label{figure:component}
\end{figure}

The input and output formats of our proposed framework are compatible with widely used packages in the field of computational biology, such as AnnData~\citep{virshup2021anndata}, Scanpy and MUON (Figure \ref{figure:datastructure_inputs}). The created graphical model is implemented using Tensorflow 2.8.0 sequential and functional APIs~\citep{abadi2016tensorflow}. Each data modality is handled by a component module with custom data preprocessor, encoder $f^{(e)}_v$, hierarchical encoder ($f^{(r)}_v$ and $f^{(b)}_v$) and decoder $f^{(d)}_v$ (Figure \ref{figure:component}). The framework also allows the user to use one component that takes multiple modalities as input, so that the component can also be used as a standalone multi-modal variational autoencoder with a shared latent space across modalities (see Figure~\ref{figure:relational_graphs}a).


Note that the decoder of view $f^{(d)}_v$ takes the sampling values from the latent representations of its parents $\hat{z}_{\mathcal{P}(v)}$ as inputs, where $\hat{z}_{\mathcal{P}(v)}$ also incorporates the latent representations of their parents $\bigcup_{i\in\mathcal{P}(v)}\mathcal{P}(i)$ (see Equation~\ref{equation:z_hat}). These dependencies may propagate back to the root (i.e. the modalities without parent). In other words, the posterior $q_\phi(z_v, c_v|x_v,b)$ would be conditioned on the latent representations of all its ancestors. To make the framework more flexible, we also allow the posterior of a specific modality to be conditioned only on the latent representations of its parents $z_{\mathcal{P}(v)}$ instead of all the ancestors $z_{\mathcal{A}(v)}$, using the $AlternativeIncorporateParents$ function:

\begin{equation*}
   AlternativeIncorporateParents(m)=f^{(b)}_m([z_\mathcal{P};\tilde{z}_m]; \theta^{(b)}_m)
\label{equation:z_hat}
\end{equation*}

To facilitate the future development and broaden the applications of this framework, we further aim to provide high-level APIs as command line tools for researchers, mid-level APIs that allow programmers to customize the model with parameters in the constructor functions, and low-level APIs for developers who wish to design their own data distribution, preprocessing steps, loss function or architecture of the neural network. The default neural network architecture used in the analysis presented in this study is shown in Table \ref{table:architecturee}.

\begin{table}[h]
  \caption{Default neural network architecture}
  \label{table:architecturee}
  \begin{tabularx}{\textwidth}{lll}
    \toprule
    Name         & Shape of Trainable Weights & Activation Function  \\
    \midrule
    $f^{(e)}_m$ & $\dim(x_m)\times 1024$ & linear\\
    & $1024 \times 512$ & Swish~\citep{ramachandran2017swish}, layer normalization~\citep{ba2016layer}\\
    & $512 \times 256$ & Swish, layer normalization\\
    & $256 \times 128$ & Swish, layer normalization\\
    & $128 \times (2\times\dim(z_m))$ & $\mu$ (linear), $\sigma^2$ (softplus)\\
    $f^{(b)}_m$  & $\dim([\hat{z}_\mathcal{P};\tilde{z}_m]) \times \dim(z_m)$ & linear \\
    $f^{(r)}_m$  & $\dim(z_m) \times \dim(z_m)$ & linear\\
    $f^{(d)}_m$  & $\dim([z_m;b])\times128$ & Swish, layer normalization \\
    & $128\times256$ & Swish, layer normalization\\
    & $256\times512$ & Swish, layer normalization\\
    & $512\times(\dim(x_m)\times$ number of parameters$)$ & custom \textsuperscript{1}\\
    \bottomrule
  \end{tabularx}
  \begin{tablenotes}
   \item[*] \textsuperscript{1} with specific activation functions according to the properties of parameters.
  \end{tablenotes}
\end{table}

\section{Training Strategy}
\label{section_a:training_strategy}
Previous studies have shown that \textit{progressive training} improves the ability of the model to learn disentangled representations across layers in hierarchical variational autoencoders~\citep{falck2021multi, li2020progressive}. We found that this is particularly helpful to prevent the offspring modality from learning redundant representations that have been encoded in the latent representations of their ancestors.

\begin{figure}[h]
  \includegraphics[width=\textwidth]{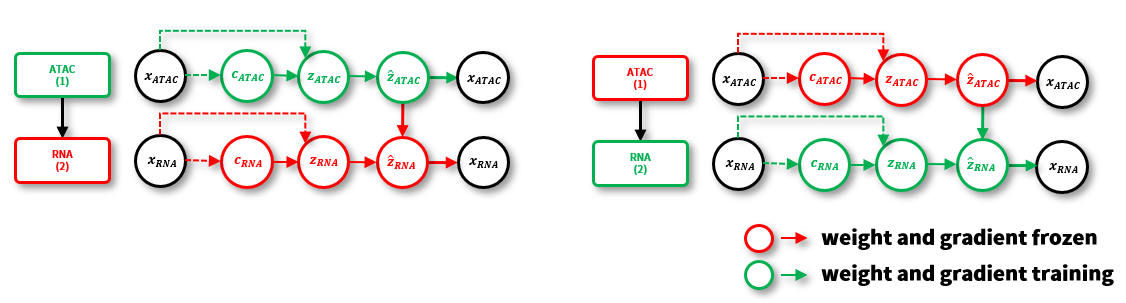}
  \centering
  \caption{Sequential training strategy. (a) the first stage and (b) the second stage of sequential training. The green box suggests the component to optimize, while the red box is the component with frozen trainable weights in each stage. The green arrows suggest regular forward and backpropagation, while the red arrows suggest the gradients of the ELBO with respect to the trainable weights in these functions that are frozen during backpropagation.}
  \label{figure:sequential_training}
\end{figure}

We apply \textit{sequential training}~\citep{wu2021greedy} based on the order of modalities (e.g. the numbers shown under the names of the components in Figure~\ref{figure:relational_graphs}) after the topological sort of the provided graph $G$. Specifically, when training the model of the $m$-th modality with its ancestors $\mathcal{A}(m)$ and their offspring $\mathcal{O}(m)$, we fix the trainable weights in $f^{(e)}_{\mathcal{A}(m)\cup\mathcal{O}(m)}, f^{(d)}_{\mathcal{A}(m)\cup\mathcal{O}(m)}, f^{(r)}_{\mathcal{A}(m)\cup\mathcal{O}(m)}$ and $f^{(b)}_{\mathcal{A}(m)\cup\mathcal{O}(m)}$. This way, the posteriors and the generative processes for the ancestor and offspring modalities remain unchanged (Figure~\ref{figure:sequential_training}). This complements progressive training and prevents ancestor modalities from learning representations that are only relevant to the offspring. As shown in previous research~\citep{wu2021greedy}, sequential training also improves the training stability and reduces memory usage. 

For the analyses presented in this study, we trained the model for 750 epochs in each stage with early stopping using the Adam optimizer~\citep{kingma2014adam} with a learning rate of 1e-4. The model is trained by maximizing the evidence lower bound of the conditional data likelihood. For the set consisting of all modalities $\mathcal{V}$, the ELBO is computed as follows:

\begin{equation}
\begin{aligned}
\log{p(x_\mathcal{V}|b)}&\geq\mathbb{E}_{q_\phi(z_\mathcal{V},c_\mathcal{V}|x_\mathcal{V}, b)}[\log(\frac{p_\theta(x_\mathcal{V},z_\mathcal{V},c_\mathcal{V}| b)}{q_\phi(z_\mathcal{V},c_\mathcal{V}|x_\mathcal{V}, b)})]=\mathcal{L}(x_\mathcal{V}, b;\theta,\phi),\\
\end{aligned}
\label{equation:elbo}
\end{equation}

where $q_\phi(z_\mathcal{V}, c_\mathcal{V}|x_\mathcal{V}, b)$ is parameterized by encoder neural networks $f^{(e)}_\mathcal{V}$. For multi-facet online clustering, we applied the single-facet VaDE trick~\citep{falck2021multi, jiang2016variational} sequentially to identify the cluster assignments of each data point. 
For a more detailed derivation, see Appendices~\ref{section_a:elbo} and~\ref{section_a:clustering}. For the analyses shown in this study, the dimensionality of the latent distribution was set to $\dim(z)=20$. To ensure the model is identifiable, we set the number of components in the mixture of independent Gaussian priors to $K=2\times\dim(z)+1=41$~\citep{khemakhem2020variational, willetts2021don}. 

\newpage
\section{Supplementary Figures}

\begin{figure}[h]
  \includegraphics[width=0.9\textwidth]{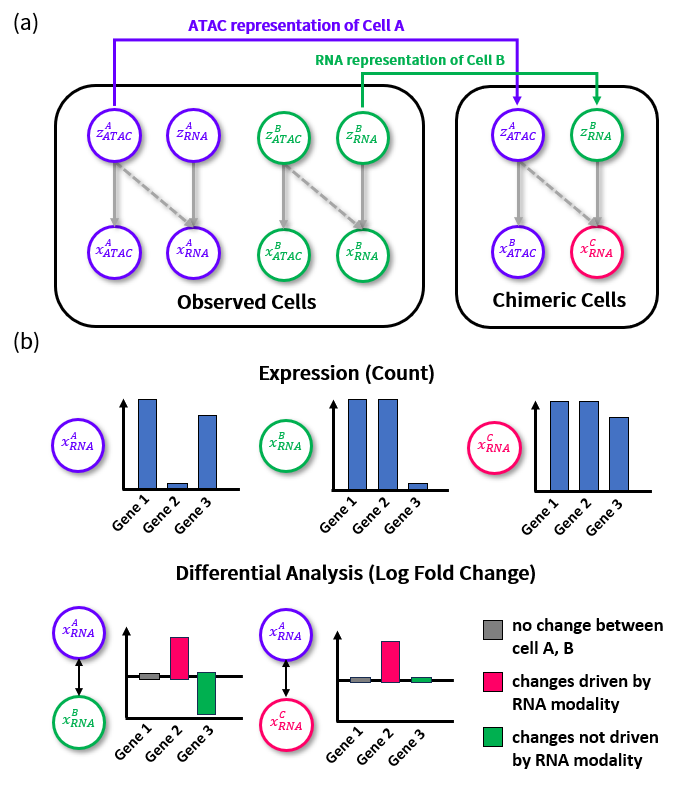}
  \centering
  \caption{Examples to decompose Bayesian differential analysis (a) the creation of the chimeric molecular profile. The different cell types are distinguished by different colors. Note that the generative process is the same across cell type A, cell type B and the chimeric cells. The molecular profiles of chimeric cells are generated using the latent representation of ATAC from cell type A, and the latent representation of RNA from cell type B. (b) Following the creation of the chimeric molecular profile, the decomposition of the Bayesian differential analysis proceeds with the comparison of differentially expressed genes from these two profiles. This allows for the identification of differentially expressed genes driven by different modalities}
  \label{figure:composite_deg}
\end{figure}

\begin{figure}[htb]
  \includegraphics[width=\textwidth]{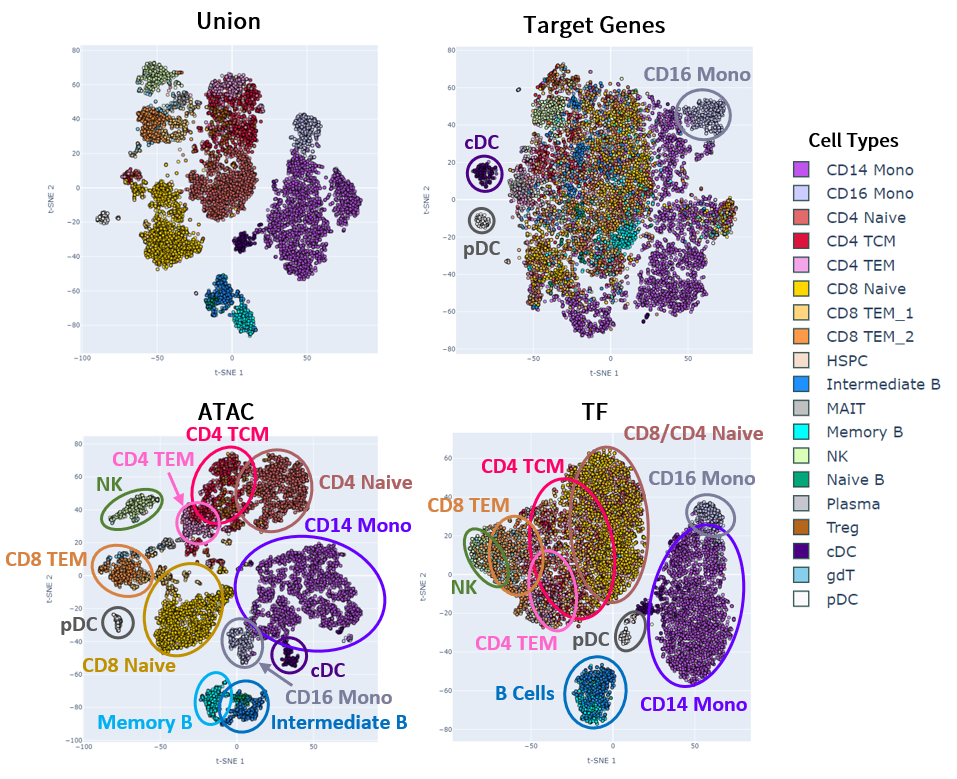}
  \centering
  \caption{Conditional latent representations for 10k PBMCs dataset. The data points are colored by the annotated cell types.}
  \label{figure_a:10x_latent}
\end{figure}


\end{document}